\documentclass{article}
\usepackage{spconf,amsmath,graphicx}
\usepackage{amssymb}

\title{HIERARCHICAL GRAPH-RNNS FOR ACTION DETECTION OF MULTIPLE ACTIVITIES}
\name{Sovan Biswas, Yaser Souri, Juergen Gall\thanks{The work has been funded by the ERC Starting Grant ARCA (677650).}}
\address{University of Bonn, Germany}
\begin{document}
%
\maketitle
\begin{abstract}

In this paper, we propose an approach that spatially localizes the activities in a video frame where each person can perform multiple activities at the same time. Our approach takes the temporal scene context as well as the relations of the actions of detected persons into account. While the temporal context is modeled by a temporal recurrent neural network (RNN), the relations of the actions are modeled by a graph RNN. Both networks are trained together and the proposed approach achieves state of the art results on the AVA dataset.	
\end{abstract}
\section{Introduction}
\label{sec:intro}

With the advent of deep neural networks and the availability of large datasets in the last decade, the performance of recent algorithms for action recognition has improved drastically. Annotating large amount of data, however, is very expensive. In particular for spatio-temporal action recognition and localization where multiple actions occur at the same time, action labels and bounding boxes would be required for each frame in order to learn a model using full supervision. The large-scale AVA 2.1 dataset~\cite{gu2018ava} for recognizing and localizing multiple actions in videos, therefore, provides only temporally sparse annotations, i.e., the persons and the actions the people perform are annotated for only one frame per second. This requires to develop methods that can be trained with such sparse annotations~\cite{gu2018ava,sun2018actor,girdhar2018better,stroud2018d3d}. 

While these works take temporal information into account, they do not model the interactions of the individual persons. While some actions of different persons are uncorrelated, other actions refer to interactions between persons like `talk to' and `listen to'. There are also actions that are often performed by several persons at the same time like `sit', `stand', `walk', or `play instrument', but there are also actions that exclude each other. For instance, if a person `drives' a car, it is very unlikely that the other persons in the car `stand' or `play an instrument'.      

\begin{figure}[t]
\centering
\begin{minipage}[h]{.9\linewidth}
  \centering
  \centerline{\includegraphics[width=\textwidth]{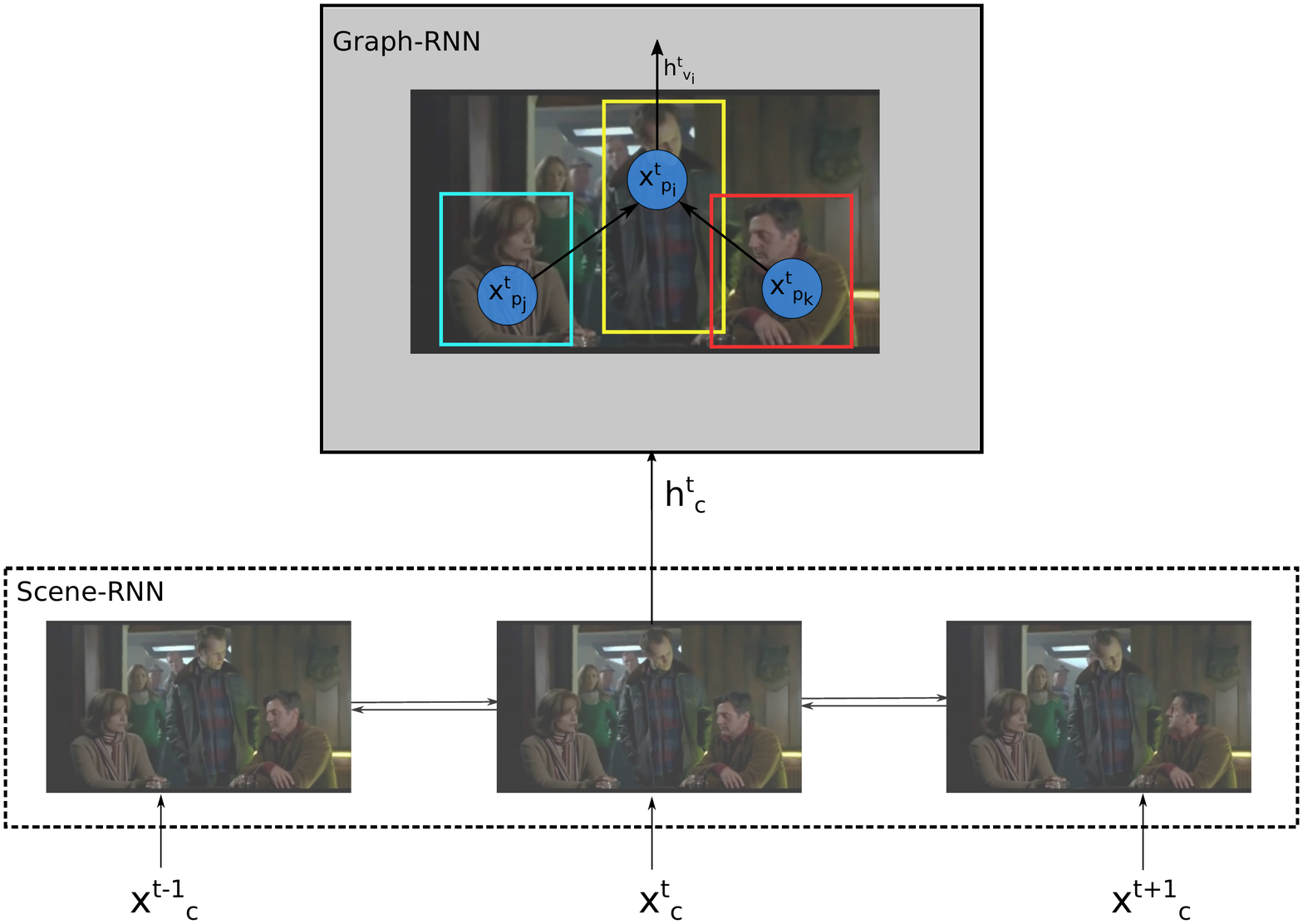}}
\end{minipage}
\caption{The proposed network jointly captures temporal context and the relations between different persons by a hierarchy. Scene RNNs, at the lower level, are used to model the temporal context of a scene. The Graph-RNN on top of it, models the relations of the actions of the detected persons.}
\label{fig:fig1}\vspace{-4.1mm}
\end{figure}                  

In this work, we therefore propose an approach that learns the relations of actions that occur at the same time and that can be learned using only sparse annotations. In order to address the sparseness of the annotations, we do not rely on tracked bounding boxes, which can be unreliable. Instead, we propose a hierarchical model that models the temporal scene context in the lower level and the relations of actions of the detected persons on the top level as illustrated in Fig.~\ref{fig:fig1}. The temporal scene context is important since the relations of the actions depend on the scene. For instance, it matters if persons are inside of a moving car or a parking car. We model the scene context by a recurrent neural network (RNN) that uses I3D features~\cite{i3d} as input. The RNN models the temporal context of the entire frames. At the top level of the hierarchy, we combine the hidden states of the scene context RNN with I3D features extracted for all detected persons in a frame. To learn the relations of the actions of all detected persons, we use a graph recurrent neural network~\cite{scarselli2009graph,gori2005new,li2016grnn}. The proposed model therefore learns the scene RNN and the graph RNN together.               

We evaluate our approach on the large-scale AVA 2.1 dataset~\cite{gu2018ava} where the approach achieves state-of-the-art results for action detection of multiple activities.  

\begin{figure*}[t]
\begin{minipage}[b]{1.0\linewidth}
  \centering
  \centerline{\includegraphics[width=1.0\textwidth]{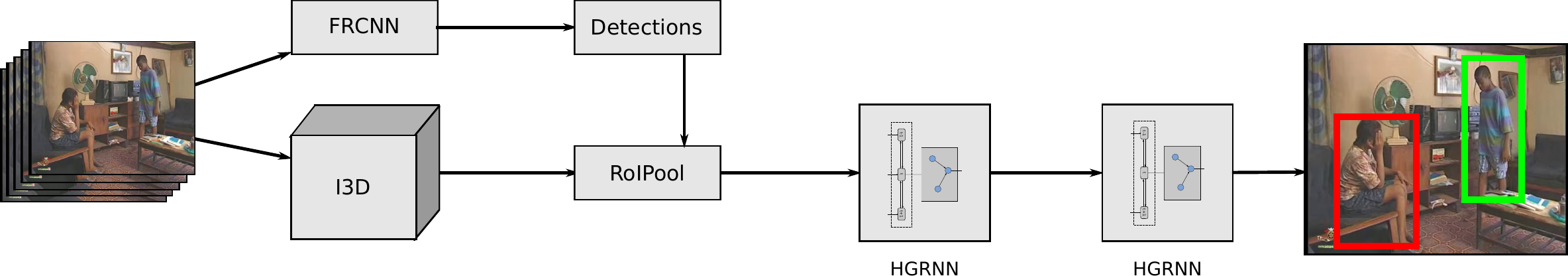}}
\end{minipage}
\caption{After detecting bounding boxes using Faster RCNN and pooling I3D features for each detected bounding box, the proposed hierarchical Graph-RNN predicts multiple class labels for each bounding box. In this example, the network detects two persons and infers the activities `sit', `watch (a person)', and `listen to (a person)' for the red bounding box and `stand', `talk to (a person)', and `watch (a person)' for the green bounding box.}
\label{fig:fig2}
\end{figure*}
	  
\section{Related Work}
\label{sec:reltdwork}
 
A common approach for action recognition and localization~\cite{gkioxari2015finding,hou2017tube,kalogeiton2017action,singh2017online} comprises the detection of the bounding boxes in each frame using object detectors~\cite{girshick2015fast,chao2018rethinking}. The detected bounding boxes are then linked to obtain action tubes, which are then classified. These approaches, however, assume that every frame is annotated. Since such dense annotations are very time-consuming, the approaches~\cite{weinzaepfel2016towards,girdhar2018better,li2018recurrent} deal with sparse annotations where the action labels and locations are annotated only for a subset of frames, e.g., each frame per second. These works, however, treat each person independently although persons tend to interact with each other.


In the context of group activity analysis~\cite{ibrahim2016hierarchical,biswas2018structural,tang2018mining}, the relations between various individual persons are used to infer the action label of the group as well as the individuals. The works~\cite{ibrahim2016hierarchical,biswas2018structural} propose hierarchical models where the individual actions are modeled at the lower level and the group activity at the top level. Our task, however, differs from group activity analysis. While~\cite{ibrahim2016hierarchical,biswas2018structural,tang2018mining} assume that each individual is part of a sports team or group and each individual performs only one action as part of a group, \cite{gu2018ava} contains multiple activities per individual and only a subset of the actions are based on interactions with other individuals.   

	
There are several types of networks that can be applied to graphs like Graph Convolutional Networks (GCN)~\cite{kipf2016semi, NIPS2015_5954} or Graph Recurrent Neural Networks (Graph RNN)~\cite{scarselli2009graph,li2016grnn,jain2016srnn}. 
These graph networks have been used in various computer vision applications such as object detection~\cite{qi2018learning}, semantic segmentation~\cite{qi20173d}, or visual question answering~\cite{teney2017graph}. For instance, \cite{yang2018graph} use an attentional GCN to model spatial relations between objects in an image. In our work, we use a Graph RNN combined with a Scene RNN to model relations between different persons as well as temporal context.             

	
\section{DETECTION OF MULTIPLE ACTIVITIES}
\label{sec:algo}
As it is defined in the AVA 2.1 dataset~\cite{gu2018ava}, the goal of the approach is to estimate for a frame the bounding boxes of all persons performing an action and for each bounding box the action labels. In contrast to other datasets, each bounding box is annotated by multiple labels. For instance, a person might `stand', `carry/hold' an object, and `listen to' another person at the same time.         


\subsection{Features}
\label{ssec:movrview}

As illustrated in Fig.~\ref{fig:fig2}, the proposed approach consists of two steps. We first detect the persons in a frame and then we use the hierarchical graph RNN to infer the action labels for each detected bounding box. For detecting the bounding boxes, we use Faster RCNN \cite{ren2015faster} with a ResNet architecture \cite{he2016deep}. The detector is fined-tuned on the dataset and the detections are performed in an action class agnostic way. 


For the temporal context of a frame $t$, we extract I3D features~\cite{i3d} for $t$ and its neighboring frames. We only consider the frames $t$ that are annotated in AVA, i.e., one frame per second, but a single I3D feature is computed over a temporal context of 33 frames. This means that the temporal receptive fields of the extracted features for the annotated frames $t$ and $t+1$ slightly overlap. Given $l$ annotated frames before and after frame $t$, we get a sequence of $L=2l+1$ I3D features
\begin{equation}\label{eq:l}
x_{c}^{t-l}, \ldots, x_{c}^{t-1}, x_{c}^t, x_{c}^{t+1}, \ldots, x_{c}^{t+l}, 
\end{equation} 
which corresponds to a temporal context of approximately $L$ seconds for AVA. 

We also use the detected bounding boxes for each sparsely sampled frame and use region pooling to extract I3D features per detected bounding box as in~\cite{gu2018ava}. This gives for each detected bounding box $i$ in frame $t$ another I3D feature vector that is denoted by $x_{p_i}^t$. We will use these features as input for the hierarchical graph RNN.

\subsection{Hierarchical Graph RNN (HGRNN)}
\label{ssec:HGRNN}
The proposed hierarchical graph RNN as illustrated in Fig.~\ref{fig:fig1} comprises two types of RNNs that are trained together. At the lower level, the scene RNN, which is described in Section~\ref{sec_srnn}, models the temporal scene context. The scene context estimated by the scene RNN is then used as input for the graph RNN, which models the relations of the actions performed by the detected persons. The graph RNN, which is described in Section~\ref{sec_grnn}, predicts then for each bounding box multiple action labels.


\subsubsection{Scene RNN}
\label{sec_srnn}
The scene RNN takes all $L$ scene features $x_{c}^t$ as well as the features extracted for each detected person $x_{p_i}^t$ in all $L$ frames as input and predicts a hidden state $h_c^t$ only for frame $t$. Since it uses the frames before and after the frame $t$, we use a bidirectional RNN with GRUs. At this stage, we do not model any relations between the persons and simply perform maxpooling over all features $x_{p_i}^t$ at each frame.        
%
\begin{eqnarray}
\label{eq:tbg}
	x_P^t & = & \textit{maxpool}_i(x_{p_i}^t)\\
	x^t & = & x_{c}^t \oplus x_P^t \nonumber \\
	h_c^t & = & \textit{biGRU}(x^t;h_c^{t-1}, h_c^{t+1}).
\end{eqnarray}
While $x_P^t$ denotes the maxpooled person feature for all detected persons in frame $t$, $\oplus$ denotes the concatenation of two vectors.

\subsubsection{Graph RNN}
\label{sec_grnn}

Given $i$ detected persons in frame $t$ and the corresponding features $x^t_{p_i}$ as well as the hidden scene representation $h_c^t$ estimated by the scene RNN, we now model the relations of the detected persons to infer the activities for each of them. To this end, we represent each detected person as a node ($v_i \in V$) of a fully connected graph. This means that we consider all possible relations how an action of a person effects the actions of the other persons. 

As in \cite{scarselli2009graph,gori2005new,li2016grnn}, we use a graph RNN that iteratively updates the hidden representation for each node $v_i$ based on the intermediate representations of the other nodes. In our case, the equations for the graph RNN are given by    
\begin{eqnarray}
\label{eq:ggru}	
	x_{v_i}^{(j)} & = & \textit{maxpool}_{v_i \in V}(h_{v_i}^{(j-1)}) \nonumber \\
	a_{v_i}^{(j)} & = & x_{p_i} \oplus h_c \oplus x_{v_i}^{(j)}  \nonumber \\
	z_{v_i}^{(j)} & = & \sigma(U^za_{v_i}^{(j)} + W^zh_{v_i}^{(j-1)}) \nonumber \\
	r_{v_i}^{(j)} & = & \sigma(U^ra_{v_i}^{(j)} + W^r{h_{v_i}}^{(j-1)}) \nonumber \\
	s & = & tanh(U^sa_{v_i}^{(j)} + W^s{(h_{v_i}^{(j-1)} \circ r_{v_i}^{(j)})}) \nonumber \\
	h_{v_i}^{(j)} & = & (1-z_{v_i}^{(j)})\circ h_v^{(j-1)} + z_{v_i}^{(j)} \circ s \
\end{eqnarray}
where we omitted the frame index $t$ for the ease of reading and $\circ$ denotes the Hadamard product. At each iteration $j$, the hidden representation for a detected person is given by $h_{v_i}^{(j)}$. To update the representation, we first maxpool the hidden representation over all nodes and concatenate it with the original person feature $x_{p_i}$ as well as the temporal scene context $h_c$ estimated by the scene RNN, which provides a longer temporal context for the graph RNN than the person features $x_{p_i}$. Using a GRU variant, $h_{v_i}^{(j)}$ is then updated. After a fix number of iterations, the estimated hidden representation for each detected person $h_{v_i}^{(j)}$ is then fed to a fully connected layer with sigmoid as activation function to infer all action classes that are simultaneously performed.        

The entire hierarchical graph RNN consisting of the scene RNN and the graph RNN is trained jointly using the focal loss~\cite{Lin_2017_ICCV} for multi-label classification.          
 

\subsection{Implementation Details}
\label{ssec:imple}
The person detector is initialized by a ResNet-101 architecture trained on ImageNet. We finetune the person detector using Adam optimizer with a variable learning rate starting from 0.00001 and dropping by 0.5 at regular intervals. This fine-tuning is done for 100K steps with an effective batch size of 20. The I3D network is pre-trained on Kinetics \cite{i3d}. Due to memory reasons, we reduce the size of the I3D features from 1024 dimensions to 256 dimensions using a fully connected layer on top of the I3D network and finetune the network on the dataset. For finetuning, we use Adam optimizer with a variable learning rate starting from 0.0005 and dropping by 0.5 at various intervals. This fine-tuning is done for 50K steps with an effective batch size of 50. To make our model robust, we perform data augmentation using random flips and crops as in~\cite{gu2018ava}.  
 
The hierarchical graph RNN is randomly initialized and trained from scratch using 1000 steps with a batch size of 50 using Adam optimizer with a constant learning rate of 0.0001. Furthermore, we use the focal loss \cite{Lin_2017_ICCV} with $\gamma=2$. We train the network on the annotated ground-truth bounding boxes. For inference, we use the detected bounding boxes and we perform a simple multiplication of the person detection confidence with the corresponding action prediction score to obtain the final class predictions for each detection. Finally, class specific non-maximum suppression is used to remove duplicate detections.


\section{Experiments}
\label{sec:exp}

For evaluation, we use the AVA 2.1 dataset \cite{gu2018ava}. It contains 60 action classes across 235 videos of 15 minutes each for training and 64 videos of the same length for the validation set, which we use for evaluation. Sparse annotations in form of action labels and bounding boxes are provided for a single frame every second. The evaluation is performed using frame-level mean average precision (frame-AP) at IoU threshold 0.5, as described in \cite{gu2018ava}.


\begin{table}[t]
\centering
\begin{tabular}{|c|c|c|c|}
\hline
\textbf{Temporal context $l$}&\textbf{0}&\textbf{1}&\textbf{3}\\
\hline
mAP & 19.0\% & 19.5\% & 20.9\%  \\
\hline
\end{tabular}
\caption{Impact of the temporal context $l$.}
\label{tab:sliding window}
\end{table}

\textbf{Temporal Context}: To analyze the effect of increasing the temporal support, we evaluate various values for the temporal context $l$~\eqref{eq:l}. Since the frames are sparsely sampled, $l=3$ corresponds to a temporal context of 7 seconds while $l=0$ corresponds to 1 second. For the experiment, we use only RGB data without optical flow. The results in Tab.~\ref{tab:sliding window} show that increasing the temporal context increases the accuracy. This is due to the fact that certain actions such as `open' and `close' can be better recognized with a larger temporal receptive field.   
\begin{table}[t]
\centering
\begin{tabular}{|c|c|c|c|}
\hline
\textbf{Models}&\textbf{Graph-RNN}&\textbf{Scene-RNN}&\textbf{HGRNNs}\\
\hline
mAP & 19.0\% & 19.8\%  & 20.9\%  \\
\hline
\end{tabular}
\caption{Comparison of the Scene-RNN and Graph-RNN with the HGRNN.}
\label{tab:jointM}
\end{table}

\textbf{Joint Temporal and Interaction Modeling}: In Tab.~\ref{tab:sliding window}, the accuracy for $l=0$ corresponds to the case where only the Graph-RNN but not the Scene-RNN is used. In Tab.~\ref{tab:jointM}, we also report the accuracy if we use only the Scene-RNN but not the Graph-RNN. In both cases, the proposed Hierarchical Graph-RNN, which combines both RNNs in a single model, achieves a higher accuracy. This shows that both the temporal context as well as the interactions between the persons contribute to the action detection accuracy. 

\textbf{Number of Iterations}: As discussed in Section~\ref{sec_grnn}, the HGRNN block is iterated. The results in Tab.~\ref{tab:HGRNN} show that not many iterations are required. This can be attributed to the fact that our Graph-RNN already incorporates temporal information through the hierarchy and it requires only two iterations to update the hidden state of each person based on the hidden states of the other persons. In all other experiments, we use $2$ iterations.

\begin{table}[t]
\centering
\begin{tabular}{|c|c|c|c|}
\hline
\textbf{Iterations}&\textbf{1}&\textbf{2}&\textbf{3}\\
\hline
mAP & 20.6\% & 20.9\% & 20.8\%  \\
\hline
\end{tabular}
\caption{Impact of the number of iterations.}
\label{tab:HGRNN}
\end{table} 

\textbf{Ground Truth (GT) Bounding Boxes}: In order to understand the effect of the accuracy of the person detector on the action detection accuracy, we used ground truth bounding boxes during inference. The fine-tuned Faster RCNN person detector achieves an mAP of $89.09\%$ for detecting the annotated bounding boxes on AVA. If ground truth detections are used instead, the action detection accuracy increases by $5-6\%$ as shown in Tab.~\ref{tab:GT Results}. We also report the results if we use RGB and optical flow for computing the I3D features. The additional optical flow increases the accuracy by $2.7\%$ and $3.9\%$ for detected and GT bounding boxes, respectively. 

\begin{table}[t]
\centering
\begin{tabular}{|c|c|c|}
\hline
\textbf{Method}&\textbf{GT}&\textbf{Detected}\\
\hline
 RGB & 25.2\% & 20.9\%  \\
\hline
 RGB+Flow & 29.1\% & 23.6\%  \\
\hline
\end{tabular}
\caption{Quantitative comparison of the proposed method with ground truth bounding boxes and detected bounding boxes.}
\label{tab:GT Results}
\end{table}

\begin{table}[t]
\centering
\begin{tabular}{|c|c|c|}
\hline
\textbf{Method}&\textbf{flow}&\textbf{mAP}\\
\hline
AVA \cite{gu2018ava} & & 14.5\% \\ 
\hline
ACRN \cite{sun2018actor} & & 17.4\% \\
\hline
Better AVA \cite{girdhar2018better} & & \textbf{21.9\%} \\
\hline
 HGRNN - RGB &  & 20.9\%  \\
\hline
\hline
AVA \cite{gu2018ava} & \checkmark & 15.6\% \\
\hline
D3D \cite{stroud2018d3d} & \checkmark & 23.0\% \\
\hline
HGRNN - Flow & \checkmark & \textbf{23.6\%}  \\
\hline
\end{tabular}
\caption{Comparison of the proposed method with other state of the art methods. A \checkmark at the flow column indicates if optical flow has been used.}
\label{tab:results}
\end{table}	

\textbf{Comparison with State of the Art}: The proposed approach outperforms the approaches~\cite{gu2018ava,sun2018actor} by a large margin. Most interesting is the comparison to~\cite{gu2018ava} since it uses the same features but a vanilla I3D head for action detection. The proposed hierarchical Graph-RNN improves the accuracy by 6.4\% on RGB data and 8.0\% on RGB+Flow data. This clearly demonstrates the capability of the proposed hierarchical GRNN in comparison to the I3D head~\cite{gu2018ava}. We also compare our approach with the very recent works~\cite{girdhar2018better,stroud2018d3d}. While our approach outperforms~\cite{stroud2018d3d}, \cite{girdhar2018better} achieves a higher accuracy for RGB data. The gain of the accuracy is the use of a single Faster RCNN framework that detects the bounding boxes and the action classes together. Using the proposed hierarchical Graph-RNN within a Faster RCNN framework is therefore a future research direction to improve the accuracy further. However, it is unclear how much gain can be achieved if optical flow is used in addition since \cite{girdhar2018better} does not report any results for optical flow.


\section{Conclusion}
\label{sec:conclusion}
In this paper, we proposed hierarchical Graph Recurrent Neural Networks for recognizing and localizing multiple activities that occur at the same time. The model learns the temporal context as well as the interactions of the detected persons to recognize the actions. In our experimental evaluation, we have shown that the proposed model outperforms a temporal as well as a graph RNN and that the proposed approach achieves state of the art results on the AVA dataset.



\bibliographystyle{IEEEbib}
\bibliography{refs}

\end{document}